# Towards interpreting computer vision based on transformation invariant optimization

Chen Li[1,2],*, Jinzhe Jiang[1,2], Xin Zhang[2,3], Yaqian Zhao[1,2], Dongdong Jiang[1,2]

1. State Key Laboratory of High-end Server & Storage Technology, Beijing, China, 100085
2. Shandong Hailiang Information Technology Institutes, Jinan, China, 250014
3. State Key Laboratory of High-End Server & Storage Technology, Jinan, China, 250014
\* lichenslee@gmail.com

## Abstract

Interpreting how does deep neural networks (DNNs) make predictions is a vital field in artificial intelligence, which hinders wide applications of DNNs. Visualization of learned representations helps we humans understand the vision of DNNs. In this work, visualized images that can activate the neural network to the target classes are generated by back-propagation method. Here, rotation and scaling operations are applied to introduce the transformation invariance in the image generating process, which we find a significant improvement on visualization effect. Finally, we show some cases that such method can help us to gain insight into neural networks.

## 1. Introduction

Convolutional neural networks (CNN) [1] have become state-of-the-art models in computer vision tasks over the years [2]. CNN show quite well abilities even outperform humans [3]. However, it remains hard to understand fundamental mechanism behind. It's difficult to explain how the CNN make predictions and how to improve models accordingly. Beset by questions above, CNN are restricted in many fields, such as information security [4][5], healthcare [6], traffic safety [7][8] and so on.

Many efforts have been devoted to make CNN interpretable. As one of the widespread used methods, visualization play a part, if not essential, to interpret such models. Although it will never give a fundamental explanation that can describe models mathematically, visualization still makes a significant contribution to assist humans understanding CNN models. For example, Zeiler et al. propose the deconvolutional technique [9][10] to project the feature activations back to the input pixel space. Aided by this technique, they find artifacts caused by the architecture selection and therefore lead to state-of-the-art performance on the ImageNet benchmark at that time. Mordvintsev et al. [11] show the case that dumbbells always accompany a muscular weightlifter there to lift them by using visualization technique. They mention the network failed to distill the essence of a dumbbell. Visualization here can help correcting these training bias. Pan et al. [12] train a human-interpretable network by choosing and frozen the neurons which can be identified of kinds of strokes, which applied feature visualization technique.

Erhan et al. [13] apply gradient ascent to optimize the image space to maximize the neuron activity of the networks. The Deconvolutional Network [9][10] is addressed by Zeiler et al. to reconstruct the input of layers from its output. Simonyan et al. [14] come up with two methods: the saliency map and the numerical optimization with the L2-regularized image. They initialize the numerical optimization with the zero image, and the training set mean image is added to the result. However, this kind of optimization tend to generate an image full of high-frequency patterns, which is usually ambiguous to be understood. Researchers have addressed several methods [15][16] to penalize this noise. Yosinski et al. [17] combine several regularization methods to eliminate high frequency patterns and generate clearer visualized images. Recently, Zhang et al. [18] introduce the targeted universal adversarial perturbations by using random source images. From the universal adversarial perturbations images, authors mention the existence of object-like patterns can be noticed by taking a closer look. Unfortunately, these approaches are usually elaborate to specific cases that is hard to transfer to other situations.

In this paper, we propose a geometric transformation based optimization for generating of image space. We apply rotation, symmetry and scaling operations for the gradient ascent optimized images iteratively, to obtain geometric transformation robust input images, shown in Section 2. By using this method, one can obtain a visualized image with minor high frequency patterns without any elaborate regularization methods. Previous works [19][20] have already applied transformational operations for noise elimination, while the systematical transformation based optimization is not discussed before. Here, we propose the transformation based optimization method that can output some human-understandable visualization results, shown in Section 3.1. And different CNN models are compared by classified outputs in Section 3.2. We find a significant difference among CNN models. Finally, we show in Section 3.3 that the transformation based optimization will help us correcting certain training problems.

## 2. Method

In this section, we introduce the method of iterative optimization to generate a transformation robust visualization images. In general, the input image is transformed by certain operation, and the optimization is performed on this transformed image. Following, the transformation invariance is test on the concerned model. The process is carried out iteratively until the convergence condition achieved.

Previous versions of optimization with the zero image produced less recognizable images, while our method can give more informational visualization results.

### 2.1 Iterative Rotation Algorithm

The procedure is similar to the previous works [13][14], where the back-propagation is performed to optimize the input image, with the parameters of the concerned CNN fixed. The difference lies in the transformation operations that we introduce to the optimization process. That is, when the back-propagation optimization is finished, a rotation respect to a specific angle is performed to the input image. And the optimization is performed for another round. The process is carried out iteratively until the convergence condition achieved.

In consideration of the square shape, input images are clipped by the boundary and replenished by zero values of RGB. It is worth noting that construction of loss function is another way to implement transformation invariance [21], while we find that the iterative optimization method is more effective.

**Algorithm 1** The Transformation invariant visualization method

1. **Input:** the initial image $M_i$, the target class C, the stopping criterion S;

2. **while** S is not true, do

3.    **Optimization:** performing gradient-based optimization for $M_i$ until the confidence $q_C > q_{target}$, output $M_o$;

4.    **Transformation operation:** performing transformation operation for $M_o$, output $M_t$ as $M_i$;

5.    **Evaluation:** performing a series of transformation tests, and jumping out of the loop if the stopping criterion is satisfied;

6. **end for**

7. **Output:** The visualization image

    Figure 1 shows the preliminary visualization results of transformation invariant optimization, the model ResNet50 [22] is taken as an example. The first row re-verifies the nonsensical high-frequency patterns observed by many works [13]-[16]. The second row of figure 1 shows that when a simple series of rotation operation is applied with the rotation angle of 10° by 36 times, which means the image rotates for a complete revolution, some human-understandable visualization results appear. For instance, column 1 shows the optimization image targeted to the soccer ball, the feature of a hexagon edge can be found.

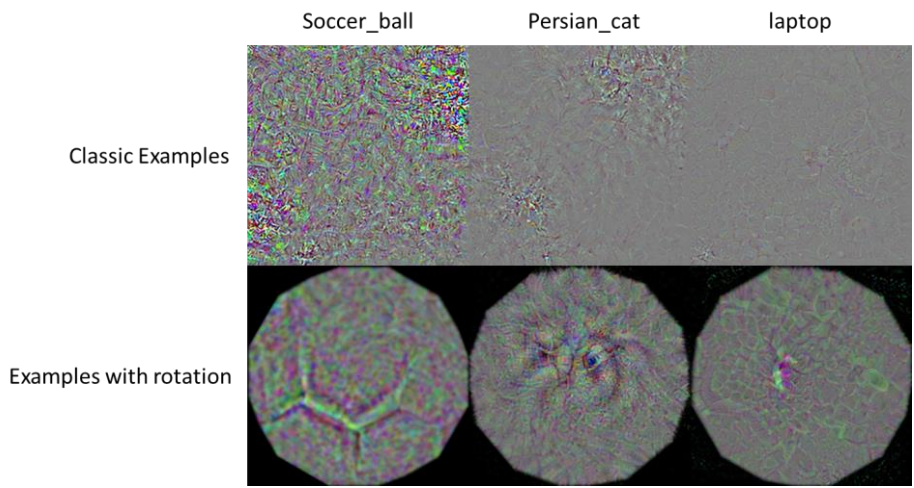

Figure 1. Visualization images for three class (soccer ball, persian cat and laptop).

First row is from the classic gradient ascent optimization, and the second row shows the result from the iterative rotational optimization.

## 2.2 Gray scale initialization

It is common to use an average-centered image as an initial start [13]-[17][20]. However, there is no evidence that an average-centered image should be the best choice as start point to optimize.

Here, we investigate the setting of the start image by analyzing the results of parallel tests. Parallel tests are performed from the initial gray scale of RGB image with (0, 0, 0), (10, 10, 10), (20, 20, 20)…(250, 250,250), (255, 255, 255), respectively. After the iterative optimization, the average gray scale change of different gray scale initial images are calculated. Figure 2 shows the result of 5 classes from ImageNet [23] in the case of ResNet50. It reveals that the different classes have a similar pattern that the RGB value around (50, 50, 50) gives a minimum change with the gray scale, while the average point (127, 127, 127) looks no particularity.

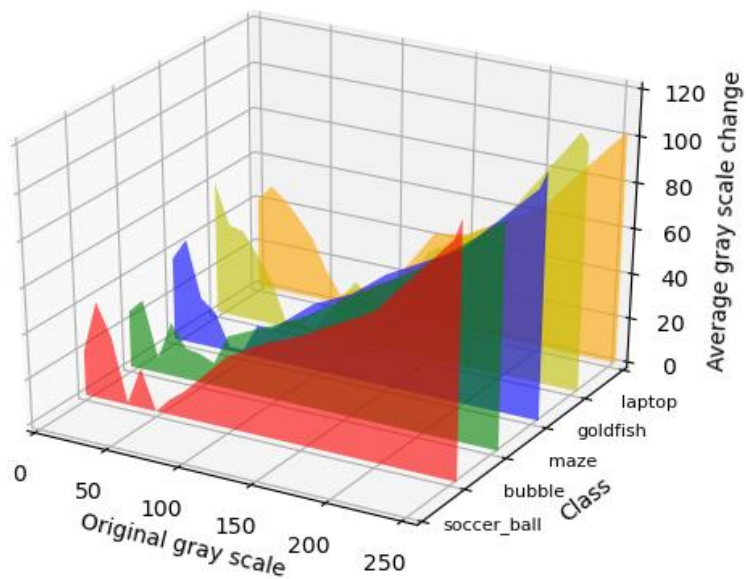

Figure 2. Average gray scale change of different gray scale initial images by ResNet50 model (Shown 5 class examples)

To further study the impact of the initial image, we adopt 2D-entropy [24]

calculation for parallel test of different gray scale images. The 2D-entropy is defined as follows:

$$H = -\sum_{i=0}^{255}\sum_{j=0}^{255} p_{ij} log_2 p_{ij}, \quad (1)$$

where $p_{ij}$ is the probability to find a pixel with gray scale of j between i, which can be counted as:

$$p_{ij} = f(i,j)/N^2, \quad (2)$$

where f(i,j) is the frequency to find the i, j couple, N is the size of the image. 2D-entropy can characterize the quantity of information, which we consider to be a good tool to quantify the effective information from the visualization.

To simplify the problem, we transfer the RGB image to a gray scale image. A BT.601-7 recommend conversion method [25] is applied:

$$V_{Gray} = int(V_R \times 0.3 + V_G \times 0.59 + V_B \times 0.11), \quad (3)$$

where $V_R$ is the value of the red channel, $V_G$ is green and $V_B$ is blue.

Figure 3 shows the parallel tests for a series of gray scale initial starts. Row (a) shows the visualization images optimized by ResNet50. It is notable that all the initial starts generate the feature of a hexagon edge, which demonstrates the robust of the transformation invariant optimization method. Furthermore, it clearly shows the difference among different gray scale initial starts. By manual picking, the gray scale between (50, 50, 50) and (60, 60, 60) are supposed to be the best choice region as the start point of optimization, because of the multi-edge structure. Row (b) shows the gray scale image transferred from row (a) by eq. (3), which reveals the key features of RGB visualization results are persisted. The 2D-entropy hot map calculated from row (b) by eq. (1) is shown in row (c). It presents that with the increase of the gray scale, the 2D-entropy decrease, except the white initial image case (255, 255, 255). This analysis is discrepant with our manual picking, which indicates the 2D-entropy can't help us filtrating good initial starts. But when scrutinizes row (c) carefully, one can find that the gray scale around (60, 60, 60) has a more intricate contour line. Instead of using row (c), we adopt second order entropy of 2D-entropy as indication of choosing initial start, shown in figure 3(d). Here, second order entropy means we calculate the 2D-entropy of the figures of 2D-entropy (row (c)). It shows obviously in row (d) that the second entropy can quantify the effective information from the visualization.

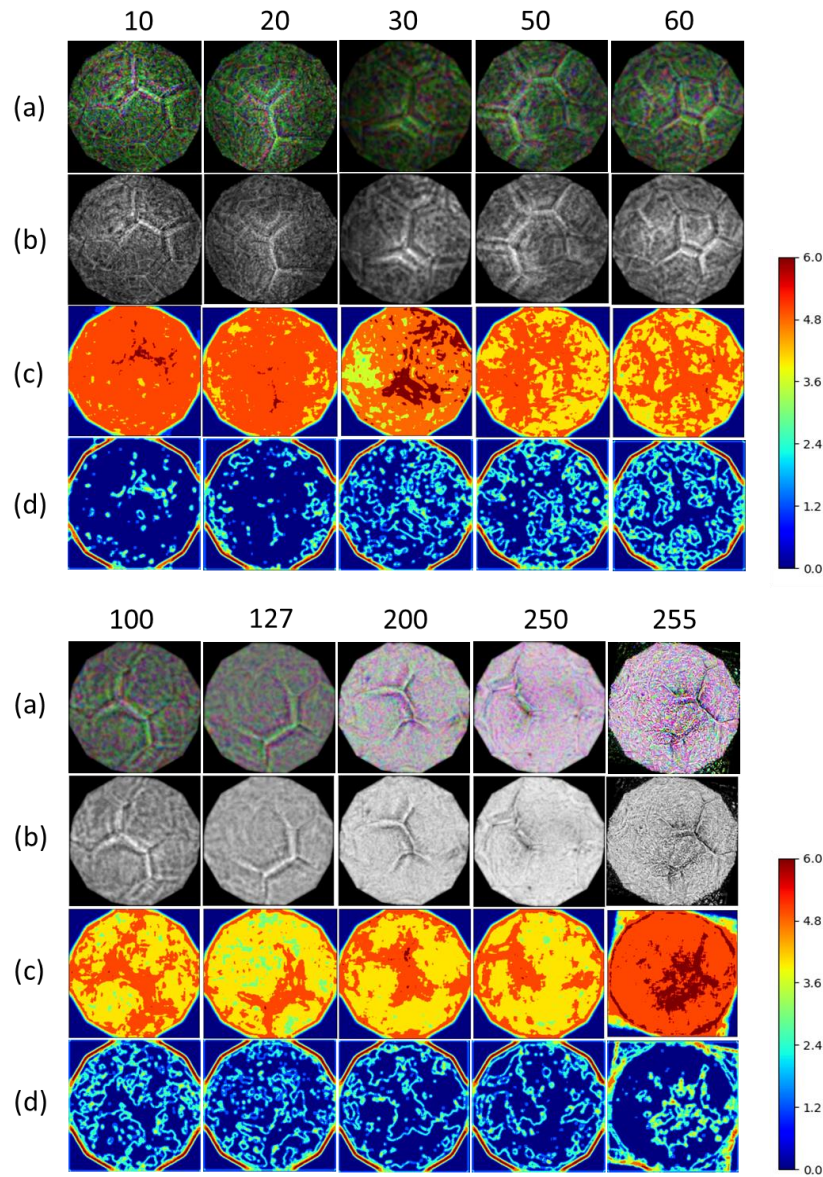

Figure 3. Visualization of soccer ball optimized by rotation. Different values are corresponding to the gray scale of initial images. Row (a) is the optimization results from different gray scale of initial images, row (b) is the RGB to Grayscale Conversion according the method of Recommendation of ITU-R BT.601-7 [25], row (c) is the 2D-entropy calculated from (b) and row (d) is the second order entropy

In order to visualize the result of Figure 3(d), a summation of second order entropy is shown in Figure 4. From the profile, we find the maximum value of the second order total entropy is around (60, 60, 60) exactly. By using this method, one can find the best initial start for any models who is interested in. In addition, we also notice a significant rise appears in the black initial image case (0, 0, 0), also shown in Figure 4. This high value of second order entropy in black initial image may originate from the zero boundary, where the pixels can only increase their gray scale value, any negative gradients will be eliminated by clipping operation. In this condition, we suppose the optimization will reduce the production of the high frequency patterns, and only the structural information is retained.

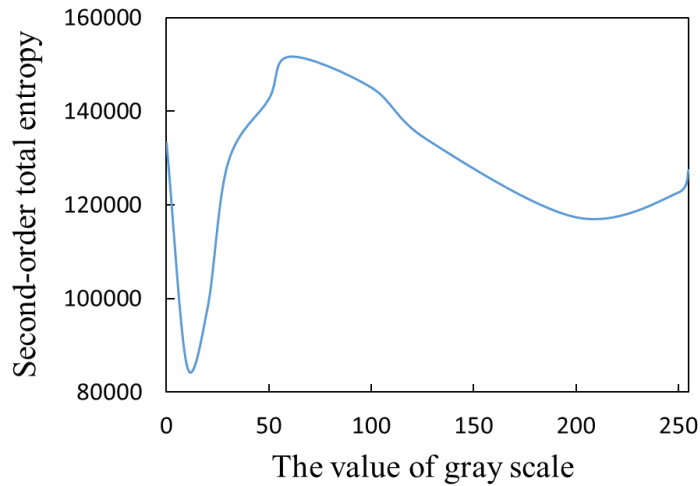

Figure 4. The second-order total entropy summation from Fig 3(d) respect to the different initial start of gray scale

## 3. Experiments

In this section, we show the cases of the visualization results by transformation based optimization. Also, different CNN models are compared by classified outputs in Section 3.2. We find a significant difference among CNN models. Finally, representative examples are shown that the transformation based optimization will help us finding certain training problems in some models. As we demonstrate in Section 2.2, models have their own best initial start. Here, in order to compare different models fairly, we adopt black background as initial start for all the cases, referring to the hypothesis we find in Section 2.2.

## 3.1 Visualization of different classes

Our visualization experiments are carried out using ResNet50 as an example. An iterative rotation optimization was applied to generate visualization images. It should be noted that although the black initial start shows a high value of second order entropy, it doesn't guarantee a human visible image. To visualize the subtle nuance of the image, we apply a color inversion:

$$P_{new} = (255, 255, 255) - P_{ori}, \qquad (4)$$

where P is a pixel of the image and P traverses all the pixels in the image. After this inversion, ResNet will not recognize the image as target class anymore, while it is much more visible for human.

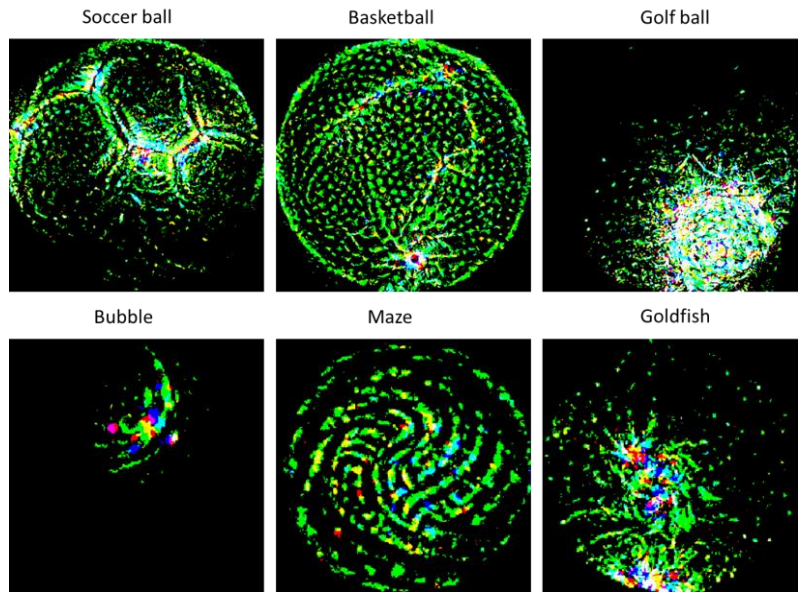

Figure 4. Visualization by iterative rotation optimization of different classes. Here the case of ResNet50 with 6 classes (Soccer ball, basketball, golf ball, bubble, maze and goldfish) is shown

Figure 4 shows the visualization results generated by ResNet50. It is impressive to see a hexagon like structure of the soccer ball class. Also the texture of the basketball, the pits of the golf ball, the path of the maze, and so on.

## 3.2 Comparison of different CNN models

Another interesting question is that how different the generated images by various models are. Figure 5 shows the comparison of ResNet50, VGG19 [26] and InceptionV3 [27] with several classes. There is no doubt that different models have different vision of things. ResNet50 seems to tend to make a freehand drawing. VGG19 prefers a colorful painting. While InceptionV3 is apt to use dash line strokes.

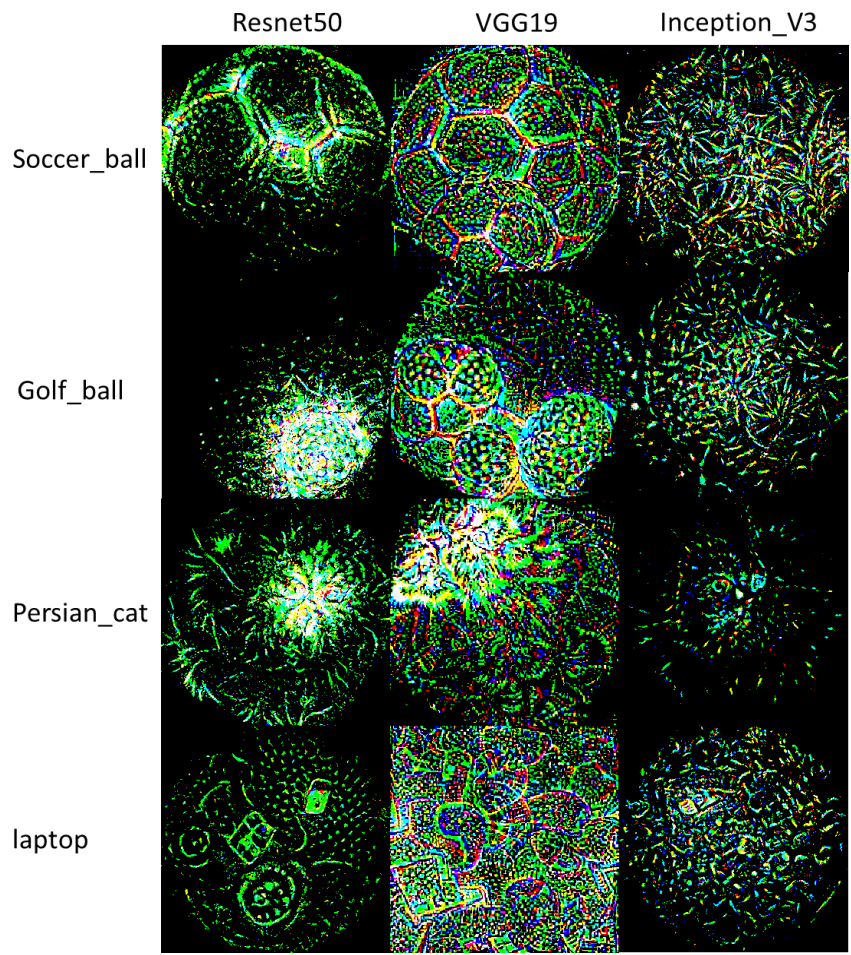

Figure 5. Visualization by iterative rotation optimization of different CNN models. Here the cases of ResNet50, VGG19 and InceptionV3 with 3 classes (Soccer ball, Persian cat and laptop) is shown

## 3.3 Debugging of models according visualization

Visualization is inadequate to make CNN models fully interpretable, but capable to assist humans understanding models. By performing transformation invariant optimization, many valuable features are found. We will show a case study of balls in this section.

In the case of soccer ball from Figure 5, both ResNet50 and VGG19 generate a hexagon structure, while InceptionV3 output an unexplainable texture. So does the golf ball case, ResNet50 and VGG19 draw the pits of the golf ball in the image, while InceptionV3 draws the elusive dash lines. When look through two images output from InceptionV3, it remind one with the grass. Coincidentally, both the football field and the green of golf grow grass. So we suppose that a CNN has a chance to combine the feature of grass with ball games.

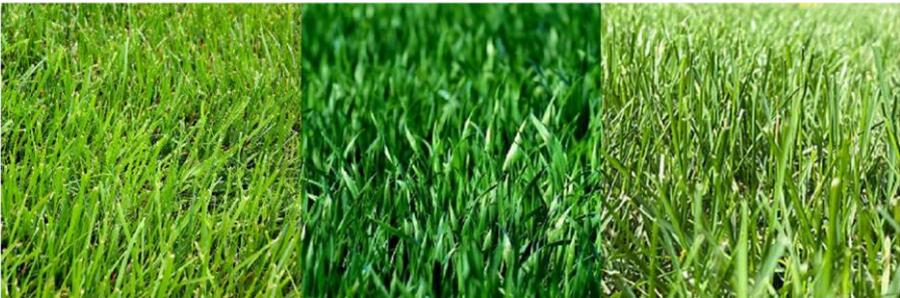

Figure 6. Prediction results of three examples of the grass by ResNet50 (R), VGG19 (V) and InceptionV3 (I)

To verify the hypothesis, some pictures of grass are sampled, and then we apply different models to classify them. The result is shown in Figure 6. On account of that there is no grass class in ImageNet [23], CNN is supposed to misclassify the image in low confidence. However, InceptionV3 classify almost all the images to balls, especially prefer golf ball as the top 1 probable class. For ResNet50 and VGG19, they prefer group grass images into animals, who frequently appear grass surroundings.

To further consolidate our hypothesis, another test is performed. First, some images of soccer ball are sampled. Following, balls in images are screened of the zero square, shown in Figure 7(2). Here, zero square means 0 input for CNN models. In the cases of

ResNet50 and VGG19, zero square is just the black image. While for InceptionV3, the range of RGB is normalized in [-1,1] [27], which means the zero square is gray scale with (127, 127, 127). Then, the screened images are got through to CNN models, and the prediction of models are shown in Figure 7. For the original images of balls, all three models make correct classifications, see the Figure 7 in label (1). However, it is consistent with the previous test after screening of balls. For example in Figure 7(a), before screening, ResNet50 has 100% confidence to classify it to soccer ball. After screening, it regards the image as a fountain with 5.124% confidence. Similar case is found for VGG19, who recognizes the screened image 7(a) as a hare. However, InceptionV3 believes it is a golf ball even the soccer ball in original image is screened. In majority cases, InceptionV3 shows a high tendency to classify the images with grass into ball group.

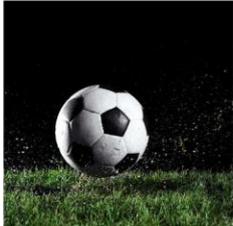

Figure 7. Prediction results of three examples with screening of the zero square by ResNet50 (R), VGG19 (V) and InceptionV3 (I). Row (o) is the original images of balls, (s) means the balls with zero screening. And the "percentage: object" pattern refer to the prediction confidence and the classification.

It should be noted that InceptionV3 can classify balls correctly in most cases. And it is robust in many other tests, which reveals further works are needed to make the model explainable. This test is try to show that our visualization method can inspire us what perhaps the model learned.

## 4. Conclusions

Regularization and frequency penalization are popular methods directly targets the high frequency noise, but these approaches somehow remove detail features sometimes. Instead, we propose a transformation invariant optimization method. The main idea is to optimize and transform an image iteratively, until the image is robust with the transformation operation. In this paper, we show that the method can generate some human understandable structures. When different models are compared, we find some valuable features to show what perhaps the models learned. And these inspire us to understand and improve models. This work is try to show that the transformation invariant based optimization make a small step toward the interpretability.

## Acknowledgements

This work was supported by National Key R&D Program of China under grant no. 2017YFB1001700, and Natural Science Foundation of Shandong Province of China under grant no. ZR2018BF011.